\documentclass{article}

\usepackage[final]{neurips_2021}

\usepackage[utf8]{inputenc} 
\usepackage[T1]{fontenc}    
\usepackage{hyperref}       
\usepackage{url}            
\usepackage{booktabs}       
\usepackage{amsfonts}       
\usepackage{nicefrac}       
\usepackage{microtype}      
\usepackage{xcolor}         
\usepackage[export]{adjustbox}
\usepackage{amsmath}
\usepackage{multirow}
\usepackage{graphicx}

\DeclareMathOperator*{\argmin}{arg\,min}

\title{Augment \& Valuate : A Data Enhancement Pipeline \\for Data-Centric AI}

\author{%
  Youngjune Lee\textsuperscript{\rm 1}, Oh Joon Kwon\textsuperscript{\rm 2}, Haeju Lee\textsuperscript{\rm 2}, \\
  \textbf{Joonyoung Kim\textsuperscript{\rm 3}, Kangwook Lee\textsuperscript{\rm 3}, Kee-Eung Kim\textsuperscript{\rm 1, 2}} \\
  {\textsuperscript{\rm 1}{School of Computing}, KAIST, Daejeon, Republic of Korea}\\
  {\textsuperscript{\rm 2}Kim Jaechul Graduate School of AI, KAIST, Daejeon, Republic of Korea}\\
  {\textsuperscript{\rm 3}Samsung Research, Republic of Korea}\\
  \texttt{dudwns511@kaist.ac.kr, \{ojkwon, hjlee\}@ai.kaist.ac.kr,}\\ \texttt{\{joon0.kim, kw.brian.lee\}@samsung.com, kekim@kaist.ac.kr}\\
}

\usepackage{graphicx}
\usepackage{subcaption}
\begin{document}

\maketitle

\begin{abstract}
  Data scarcity and noise are important issues in industrial applications of machine learning. However, it is often challenging to devise a scalable and generalized approach to address the fundamental distributional and semantic properties of dataset with black box models. For this reason, data-centric approaches are crucial for the automation of machine learning operation pipeline. In order to serve as the basis for this automation, we suggest a domain-agnostic pipeline for refining the quality of data in image classification problems. This pipeline contains data valuation, cleansing, and augmentation. With an appropriate combination of these methods, we could achieve 84.711\% test accuracy (ranked \#6, Honorable Mention in the Most Innovative) in the Data-Centric AI competition only with the provided dataset.
\end{abstract}

\vspace{-5mm}

\section{Introduction}
Data scarcity and noise are important issues in industrial applications of machine learning. These issues have been discussed a lot in the past and there are various studies. To deal with scarcity of data, methods such as auto-augmentation (\cite{autoaugment,fastautoaug,fasterautoaug}), GAN-based augmentation (\cite{DAGAN}), few-shot learning (\cite{maml,closerfewshot}) and various other methods are being developed.

Another direction of research aims to improve data quality by focusing on learning objective or model architecture such as training with noisy data  (\cite{noisetransition, lee2019robust}) and biased data  (\cite{unbiased,ovaisi2020correcting, propensity}). Moreover, there are practical methods that focus on data quality for data engineering, such as reinforcement learning based (\cite{dvrl}) and influence function based data valuation (\cite{hydra, IF}). Still, researches on data do not gain much attention compared to the evolution of the model architectures and algorithms. 

Data valuation, augmentation and representation learning (\cite{simclr,simsiam}) are known to train a robust model.
We consider this from a different point of view and intend to use these approaches to construct a pipeline that can improve the data quality for data-centric AI. We propose this pipeline, and achieved 84.711\% test accuracy (ranked \#6 and won an Honorable Mention) in Data-Centric AI competition only with the provided training data and an appropriate combination of previously mentioned methods.

\section{Competition description}
\begin{figure}
\centering
\captionsetup{justification=centering}
  \begin{subfigure}{0.21\textwidth}
  \centering
    \includegraphics[width=18mm,height=0.055\textheight]{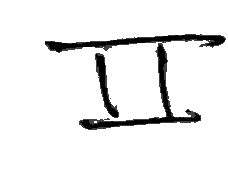}
    \caption{} \label{fig:1a}
  \end{subfigure}
  \begin{subfigure}{0.21\textwidth}
  \centering
    \includegraphics[width=18mm,height=0.055\textheight]{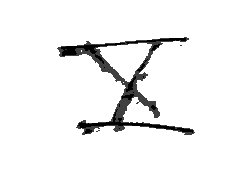}
    \caption{} \label{fig:1b}
  \end{subfigure}
  \begin{subfigure}{0.21\textwidth}
  \centering
    \includegraphics[width=18mm,height=0.06\textheight]{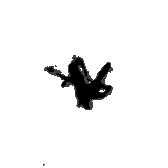}
    \caption{} \label{fig:1c}
  \end{subfigure}
  \begin{subfigure}{0.21\textwidth}
  \centering
    \includegraphics[width=18mm,height=0.06\textheight]{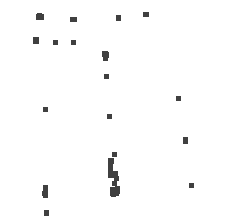}
    \caption{} \label{fig:1c}
      \end{subfigure}
  \caption{Example of noisy data. (a), (b) are mislabeled. \\ (a) GT(Ground Truth): ii, LB(Label): i. (b) GT: x, LB: ii. (c), (d) are noisy pictures.}
  \label{fig:figure_perturb}
  \vspace{-0.4cm}
\end{figure}

The Data-Centric AI competition setting aims to improve dataset given a fixed model architecture, which is the opposite of other competitions that aims to find a suitable model (algorithm) for a given dataset. The fixed model is a modified ResNet-50 (\cite{resnet}) whose input is of size 32x32. The dataset has train-validation splits, which totals to roughly 3,000 hand-written Roman numerals. The train split is unbalanced, and some data are perturbed by noise and wrong labels as in Figure \ref{fig:figure_perturb}. Another split called ``label book" consists of 52 samples that are representative of each class and can be used in place of a small test set. The model weights are selected with the best validation accuracy in 100 epochs. 
At the time of the competition, the actual test dataset was not made public. In this setting, we have to somehow construct no more than 10,000 train-validation samples from the original dataset. The goal of the competition is to improve test accuracy with the new dataset under fixed model constraints.\footnote{More details in https://worksheets.codalab.org/worksheets/0x7a8721f11e61436e93ac8f76da83f0e6} 
In other words, we need to construct a new dataset that can help the model generalize better.

\section{Methodology}
Although we focus on the competition, we concentrated on domain-agnostic techniques and relied only on the given data (i.e. we do not consider generative or collection technique) to be applied to more general tasks.

First, we considered given dataset as training data and label book as "clean" validation data. After examining the data, we identified the following challenges: (1) many training data points were perturbed by noise in input and label (2) scarcity of training data (3) imbalance of training data (4) scarcity of “clean” validation data, i.e. label book. We organized components to solve these problems. 

We define a training data point $z_i^t := (x_i^t, y_i^t) \in \mathcal{X} \times \mathcal{Y}$, where $\mathcal{X}$ is the input space and $\mathcal{Y}$ is the label space. Denote the training set as $\mathcal{D}^t := \{z_i^t\}_{i \in [|\mathcal{D}^t|]}$ and validation set with data point $z_j^v$ as $\mathcal{D}^v := \{z_j^v\}_{j \in [|\mathcal{D}^v|]}$. We denote per-sample loss given model parameter $\theta$ as $\ell(z; \theta)$. The details of our method are as follows.

\subsection{Overall pipeline}
The overall pipeline to solve the previous problems is as follows. First, we computed data value or influence(Section \ref{sub:clean_value}) for deleting negatively affecting data on validation loss. Then, we applied data augmentation(Section \ref{sub:aug}). However, this augmented dataset still had mislabeled and noisy data because the initial cleansing did not fully remove the undesirable points. Hence, we trained the model using contrastive learning and conduct secondary cleansing(Section \ref{sub:clean_cont}) from there. For this part, we set the image size to 64x64 in order not to lose too much information on the image.

\begin{figure}
\centering
    \includegraphics[width=120mm]{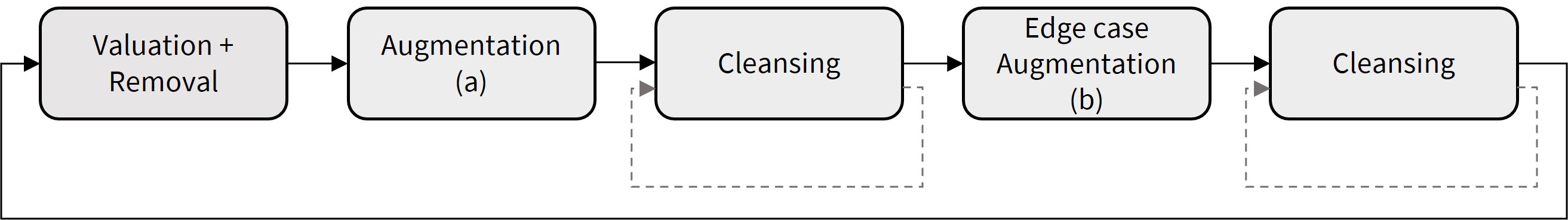}
    \caption{Our overall pipeline for get new dataset. First, we remove negative influence data after which we apply Faster AutoAugment as in (a). Next, we conduct contrastive learning for cleansing. The dashed line means several iterations. (b) is concentrated augmentation for edge case. Finally, do one more cleansing operation. We repeat this process for get full training dataset.}
    \label{fig:pipeline}
    \vspace{-0.2cm}
\end{figure}

In addition, we concentrate on some edge cases and apply more data augmentation for a more compact distribution of data representation. Finally, we apply cleansing again. We alternated cleansing and augmentation of this pipeline for a number of times to obtain a full training data. Our final pipeline is in Figure \ref{fig:pipeline}. Moreover, we used the given model architecture from the competition to provide the same inductive bias on the dataset.

\subsection{Data cleansing via data valuation} \label{sub:clean_value}
 We began by addressing training data points that were negatively impacting the model generalization performance. Rather than manually cleaning the data, our methodology focused on utilizing influence function (IF) (\cite{IF}). The idea behind IF is to ``upweight" a training point to measure its influence at inference. We define an upweighted loss on the $i$-th training point and optimized model parameter $\theta_i^*$ as follows:
  \begin{equation}
 \label{loss_if}
     \mathcal{L}_{t}^{up}(z^t_i; \theta) = 
     \frac{1}{|\mathcal{D}^t|}\sum_{k=1}^{|\mathcal{D}^t|} \ell(z^t_k;\theta) + \varepsilon_i \ell(z^t_i;\theta), \qquad \theta_i^* := \argmin_\theta \mathcal{L}_{t}^{up} (z^t_i; \theta)
 \end{equation}
 where $\varepsilon_{i}$ is small weight on $i$-th train data point.  Based on Equation \ref{loss_if}, we compute the training data influence on the validation loss by IF:
  \begin{equation}
 \label{IF}
     \text{IF}(z^t_i, z^v_j) = -\frac{1}{|\mathcal{D}^{t}|}\frac{d\ell(z^{v}_{j}; \theta^*_i) }
     {d\varepsilon_{i}}  \mathrel{\bigg|}_{\varepsilon_{i}=0},
 \end{equation}

 which means the change in the validation loss of the $j$-th validation data point by upweighting the $i$-th training point. In order to circumvent the time complexity of computing the inverse Hessian, we employed HyDRA(\cite{hydra}) to approximate the influence via hyper-gradient. Since the validation dataset (i.e. label book) was small, we filtered out data points in a conservative manner, excluding only those with large negative influences. 
 
 For more details, we calculate the influence of each training data point for each validation data point via Equation \ref{IF}. Next, we calculate the minimum 
 (min) and standard deviation (std) of all training data influence for each validation point. And remove them if influence less than min + std. We found that the model can be overfitted to the label book if we set the influence criterion too large.

\subsection{Data cleansing via contrastive learning} \label{sub:clean_cont}
 As a second step, for deleting or relabeling obviously perturbed data in pixels and label, we employed supervised contrastive learning with Siamese network\footnote{We modified https://keras.io/examples/vision/siamese\_contrastive/} trained on the dataset obtained so far. This was trained to distinguish between data of the same class and data of another class. In order to make paired examples for contrastive learning, we applied shear, inversion, shift, rotation, zoom and Gaussian noise augmentation. 
 
\begin{figure}
\centering
  \begin{subfigure}{0.21\textwidth}
  \centering
    \includegraphics[width=34mm,height=0.14\textheight,frame]{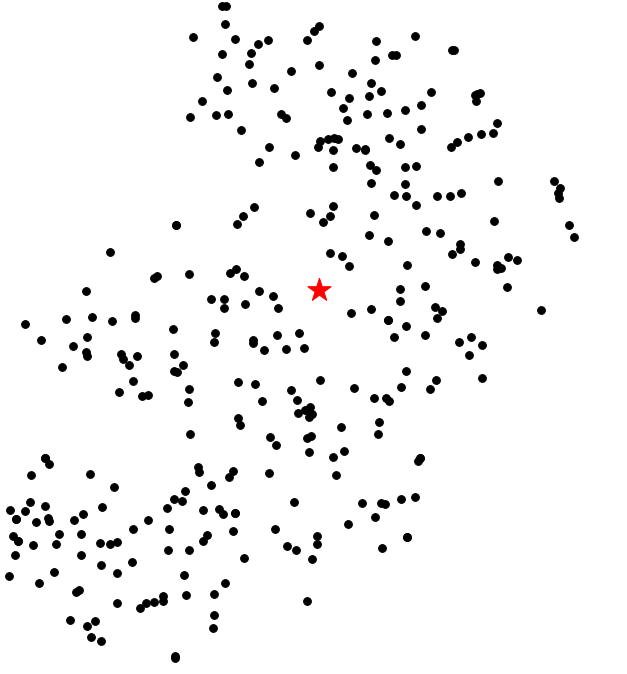}
    \caption{} \label{fig:1a}
    \end{subfigure}
  \begin{subfigure}{0.21\textwidth}
  \centering
    \includegraphics[width=3mm,height=0.14\textheight]{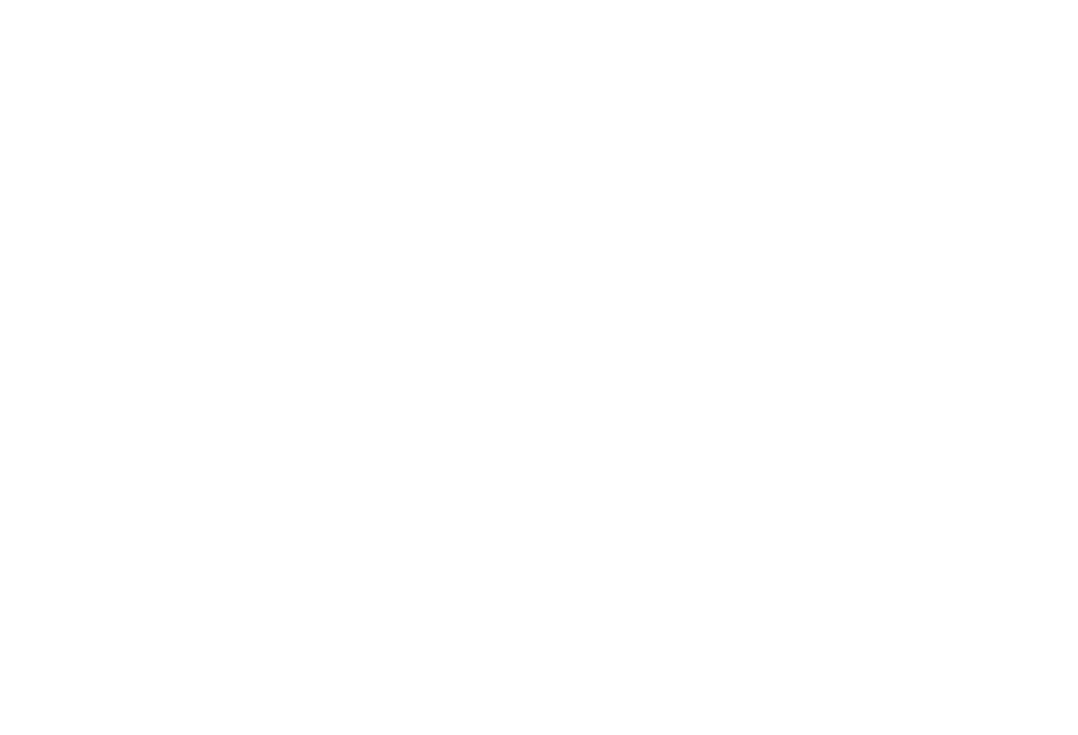}

  \end{subfigure}
  \begin{subfigure}{0.21\textwidth}
  \centering
    \includegraphics[width=34mm,height=0.14\textheight,frame]{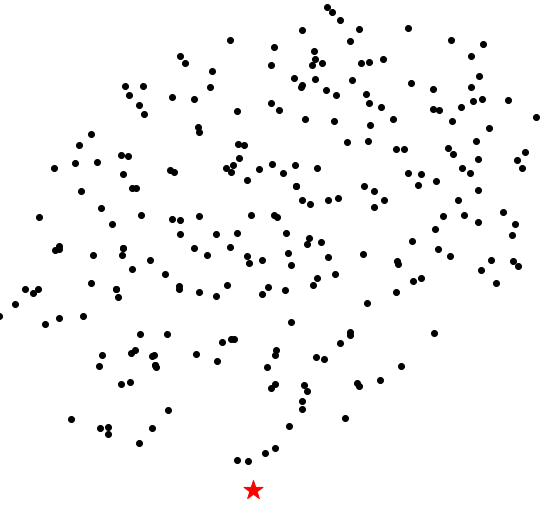}
    \caption{} \label{fig:1b}
  \end{subfigure}

  \caption{Example of visualization with learned representation. Each red star is a representative sample of a different class from the cluster. (a) shows a mislabelled point that is placed within the cluster where all its neighbors are of different labels. On the other hand, (b) shows a noisy data that lying at the boundary of the different class cluster.}
  \label{fig:figure_tsne}
  \vspace{-0.2cm}
\end{figure}

 Using the learned representation, we could visualize the dataset by projecting their feature in 2-D space via t-SNE (\cite{tsne}), and identified the data points that were obviously mislabeled or noise. Thus, using the k-nearest neighbor distances and labels, we fixed the label if it was obviously a labeling error (all neighbors having the same but a different label as in Figure \ref{fig:figure_tsne}.(a)), or dropped the data point if it didn’t obviously belong to a cluster (the closest neighbor being far from the data point or it is around a different class cluster as in Figure \ref{fig:figure_tsne}.(b)).

\subsection{Data augmentation} \label{sub:aug}

We employed a augmentation technique, Faster AutoAugment (FAA, \cite{fasterautoaug}) to address training data scarcity. Since train data is not enough, we thought that the classifier would not learn properly because the middle of the data distribution may not be filled. From this point of view, we decided to use FAA which fills in the missing data points. In order to make the augmentation fit the gray scale image setting of the competition, only shear, inversion, translation, rotation, and Gaussian noise were used for the policy search space. In this process, we balanced the number of samples in each class to solve the imbalance.

In addition, we conduct further edge case augmentations. For this, we retrain the model with augmented and cleaned dataset. After that, we project the learned penultimate features onto 2D space with t-SNE from which we identified data points that obviously belong to a cluster but far from the closest same-class neighbor as edge-case data point. On such data points, we applied further data augmentation using inversion, shift, zoom and rotation.

\section{Results}

\begin{table}[ht]
    \centering
    \resizebox{0.45\columnwidth}{!}{
    \begin{tabular}{cc}
        \toprule[1pt]
        Method  & Accuracy \\ \midrule
        Full Pipeline (ours) & \textbf{0.84711} \\
        HyDRA, FAA, Inversion & 0.82603 \\
        HyDRA, FAA & 0.80950 \\
        HyDRA, Random Augmentation & 0.75165 \\
        HyDRA & 0.67562 \\ \midrule
        Baseline & 0.64463 \\ \bottomrule[1pt]
    \end{tabular}}
    \vspace{3mm}
    \caption{Score for our pipeline component transformations}
    \label{tab:all_submission}
    \vspace{-0.3cm}
\end{table}

Table \ref{tab:all_submission} presents performance for our pipeline component transformations. We conducted the competition with subtle changes based on the our pipeline and report the accuracy on the leaderboard\footnote{Leaderboard can be found at \url{https://https-deeplearning-ai.github.io/data-centric-comp/}}.
Each of them is a combination of our components. Cleansing was included in all combinations except for the Baseline (no modification of the data). Random Augmentation means randomly augmented instead of FAA, and inversion means pixel inversion augmentation to stabilize the representation instead of augmenting edge cases. 

As a result, we were able to reconstruct data that had a good impact on performance with small and noisy dataset. At the time of submission, because the actual test set was not disclosed and limit of submission in the competition, the performance of each component was not systematically conducted and not optimized. Hence, there is a room for performance improvement. Finally, we got 84.711\% accuracy, taking the 6th place(except duplicated ranks, about 1.1\% difference from 1st place). Our approach received the Honorable Mention in the Innovative category by our pipeline.

\section{Conclusion}
Even though our approaches were not fully developed and were mostly manual at the time of submission, we confirmed the feasibility of such pipeline with the Data-Centric AI competition. For future work, we are going to improve this method with a more algorithmic approach for real-world applications on a more optimized automated pipeline toward data-centric AI.

\begin{ack}
This work was supported by the National Research Foundation (NRF) of Korea (NRF-2019R1A2C1087634), and the Ministry of Science and Information communication Technology (MSIT) of Korea (IITP No. 2020-0-00940, IITP No. 2017-0-01779 XAI and IITP No. 2019-0-00075, Artificial Intelligence Graduate School Program(KAIST))

\end{ack}



{
\small
\bibliographystyle{plainnat}
\bibliography{bibtex}

}

\end{document}